\pdfoutput=1
\documentclass[11pt]{article}

\usepackage[final]{acl}

\usepackage{times}
\usepackage{latexsym}

\usepackage[T1]{fontenc}
\usepackage[utf8]{inputenc}

\usepackage{microtype}

\usepackage{inconsolata}

\usepackage{graphicx}
\usepackage{xspace}
\usepackage{multirow}
\usepackage{booktabs}
\title{ANNA: Abstractive Text-to-Image Synthesis with Filtered News Captions}

\author{Aashish Anantha Ramakrishnan \quad Sharon X. Huang \quad Dongwon Lee \\
  The Pennsylvania State University, State College, PA, USA \\
  \texttt{\{aza6352, suh972, dongwon\}@psu.edu}
  }

\newcommand{\n}{{\sf ANNA}}

\begin{document}
\maketitle
\begin{abstract}
  Advancements in Text-to-Image synthesis over recent years have focused more on improving the quality of generated samples using datasets with descriptive prompts. However, real-world image-caption pairs present in domains such as news data do not use simple and directly descriptive prompts. With captions containing information on both the image content and underlying contextual cues, they become \emph{abstractive} in nature. In this paper, we launch {\n}, an Abstractive News captioNs dAtaset extracted from online news articles. We explore the capabilities of current Text-to-Image synthesis models to generate news domain-specific images using abstractive captions by benchmarking them on {\n}, in both zero-shot and fine-tuned settings. The generated images are judged based on visual quality and image-caption alignment with ground-truth samples. Through our experiments, we show that different fine-tuning techniques achieve limited success in understanding abstractive captions but still fail to consistently learn the relationships between syntactic and contextual features. The Dataset is available at \url{https://github.com/aashish2000/ANNA}.

\end{abstract}

\begin{figure}
      \centering
      \includegraphics[width=\columnwidth]{"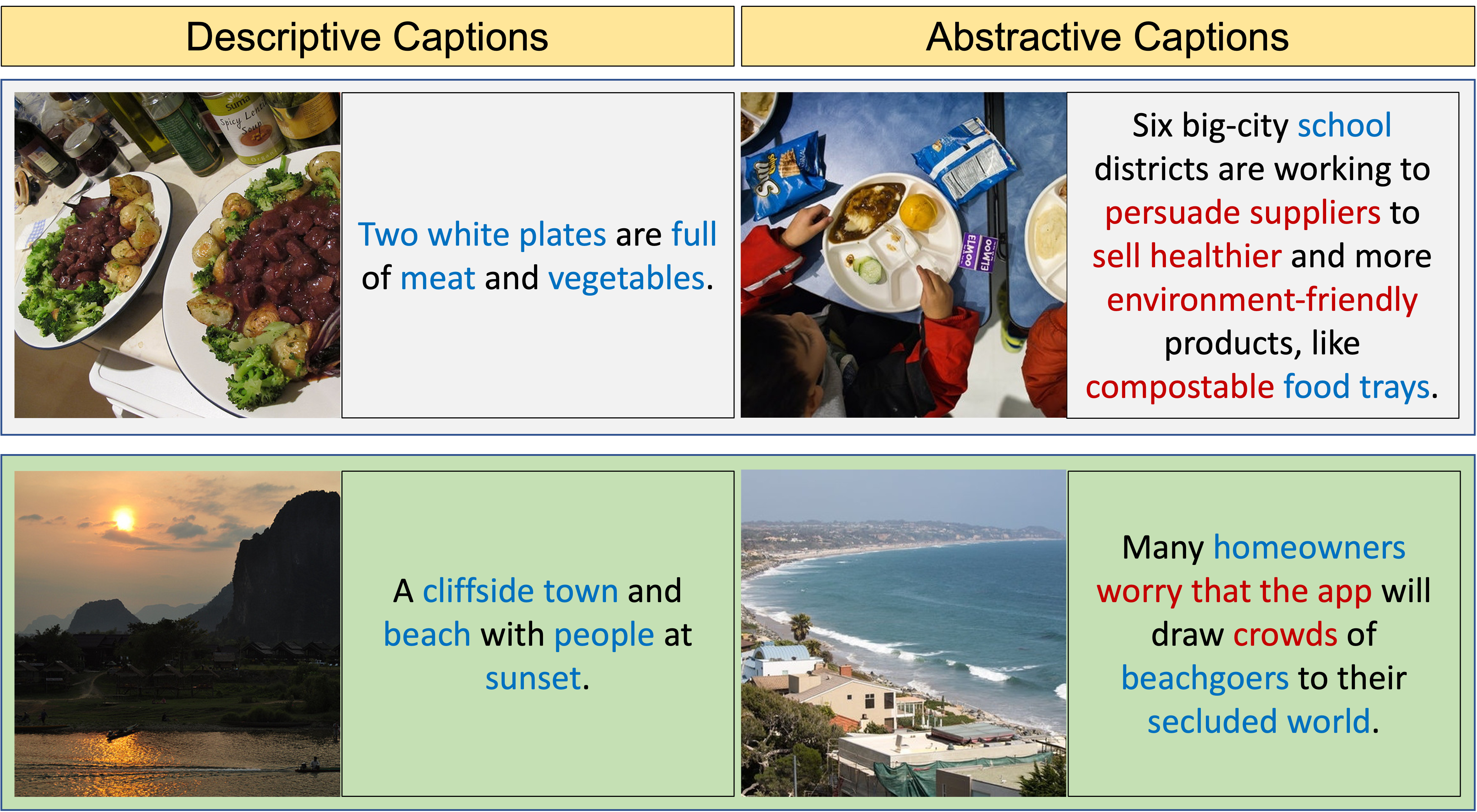"}
      
      \caption{Example of descriptive captions from the COCO Captions dataset \citep{Chen2015-qj} (Left) and abstractive captions from the {\n} (Right). Words highlighted in \textcolor{blue}{Blue} directly translate to visual entities while words highlighted in \textcolor{red}{Red} showcase contextual cues and syntactic variations present.}
      \label{fig:abstractive_captions}
  \end{figure}

\section{Introduction}
\label{section:introduction}

Image Generation has been improving by leaps and bounds over the last few years thanks to advancements in Generative Modeling approaches and the availability of higher compute capacities \citep{Tsirikoglou2020-pu}. Areas such as Text-to-Image (T2I) synthesis have grown in prominence due to the development of model pre-training paradigms on vast image-text pairs mined from the internet \citep{Schuhmann2021-er}. This has promoted the use of generators for a variety of applications such as online content creation, art synthesis \citep{Ramesh2022-kc}, and even more malicious use-cases such as DeepFake generation \citep{Zhang2022-ix}. With Internet news media and social networking websites becoming the preferred forms of information distribution, the impact that generative modeling, especially semantically-relevant image generation can have on the news media industry is quite significant.
  
Commonly, T2I synthesis has leveraged descriptive prompts, where salient visual objects are listed as keywords with instructions regarding their spatial arrangement and physical attributes. However, real-world image captions such as those found in news media are linguistically diverse to convey maximal information to readers without re-stating obvious facts \citep{Federico2016-so}. This is important in news media, where images are used to support the article's narrative and provide additional information to the reader. Non-descriptive captions make use of syntactic awareness \citep{Deacon2018-zu} and context cues to provide an efficient form of communicating situational information. Linguistic theories such as pragmatic reasoning \citep{Grice1975-mh} support the idea that the informativeness of text is influenced by its relevance to the context, especially in news captions \citep{Vedantam2017-fa}, \citep{Van_Miltenburg2016-pl} \citep{Nie2020-pk}. News captions can also vary in their level of abstraction, with some examples focusing on only a small subset of the image's visible characteristics to convey a specific message. Large Vision and Language models have shown evidence of emergent commonsense reasoning abilities to handle a variety of comprehension tasks. We select the task of news image generation from {\bf abstractive} (beyond being {\em descriptive}) captions to evaluate their ability to generate well-aligned outputs. 

Current datasets for T2I synthesis are either focused on narrow domains with simple, descriptive prompts or contain minimally filtered image-text pairs from a multitude of online sources. There are not many domain-specific datasets with image caption pairs containing situational information in addition to image descriptions. Additionally, while most models use improved visual quality of output images to be indicators of superior performance, not much focus is placed on evaluating the correlation between the output image and input text captions. This becomes more important when dealing with captions whose features are only partially aligned with the ground truth images due to their non-descriptive nature. To evaluate the task of news image synthesis, we design {\n}, a dataset containing abstractive news image-caption pairs. Abstractive captions can motivate T2I synthesis models to effectively identify these different feature types, and their relative importance and appropriately represent them in generated images. With current T2I architectures implicitly delineating between visual content and situational features, we provide detailed visualizations of both their success and failure cases on {\n} and the need for a better understanding of sentence structures for generating image features.
  
Our contributions in this paper can be summarized as the following:
  \begin{itemize}
      \item We introduce {\n}, a dataset containing approximately 30K abstractive image-caption pairs from popular media organizations
      \item We show how current open-source T2I architectures can understand abstractive captions through zero-shot and fine-tuned settings
      \item Using an exhaustive set of evaluation metrics, we benchmark popular T2I architectures based on generated image quality, image similarity to ground truth images, and human preference alignment with reference captions
  \end{itemize}

\begin{figure*}
    \centering
    \includegraphics[width=0.75\textwidth]{"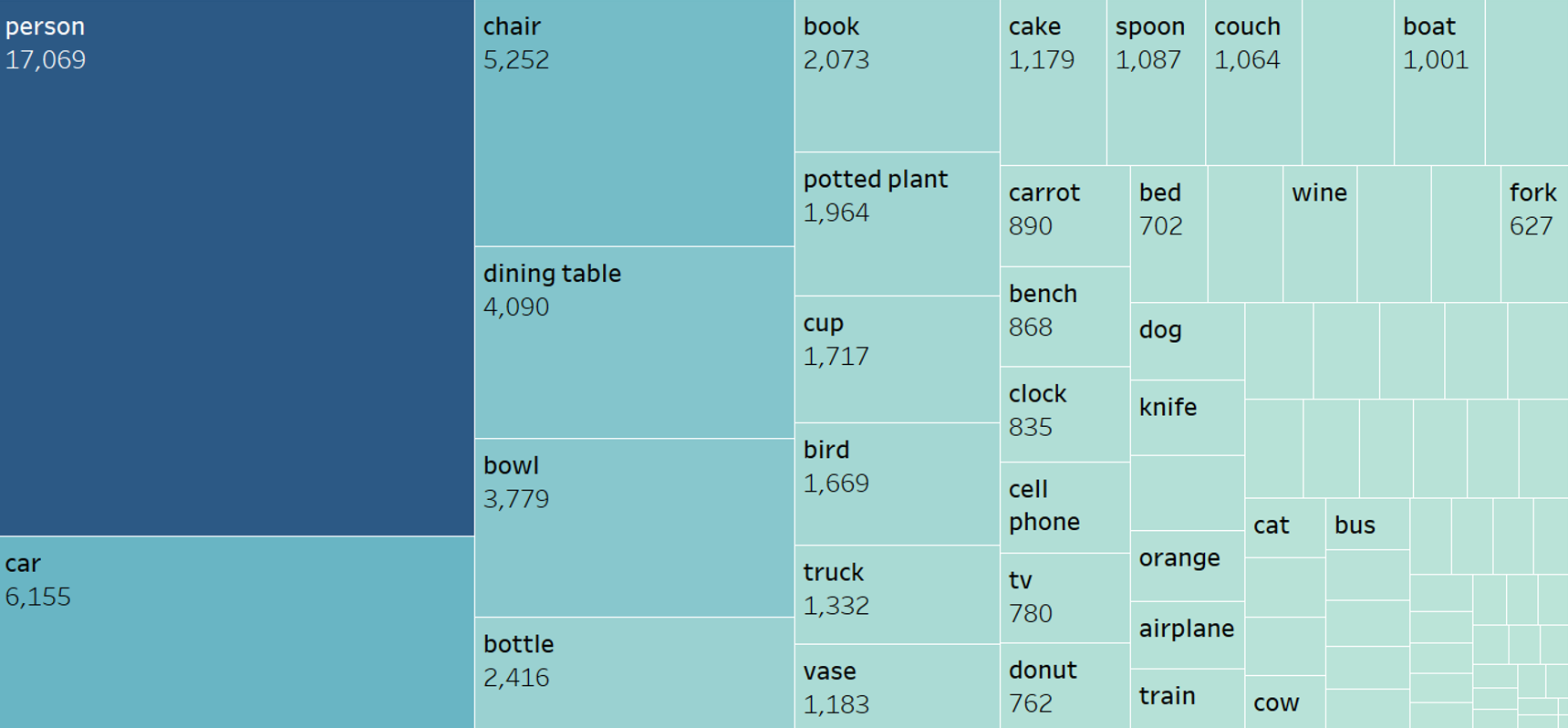"}
    \caption{Object Frequency Analysis using Treemaps}
    \label{fig:object_freq}
\end{figure*}

\section{Related Work}

\paragraph{Text-to-Image Synthesis}
The advent of Generative Adversarial Networks (GAN) \citep{Goodfellow2014-zz} sparked significant innovation in the field of T2I generation. Attention Networks were leveraged by initial GAN-based approaches for learning image-text correlations within selected datasets \citep{Zhang2017-kf}, \citep{Xu2018-rp}, \citep{Zhu2019-xs}. These models however required large amounts of task-specific data to achieve good performance. With Transformer-based architectures \citep{Vaswani2017-ju} becoming more successful at encoding visual and linguistic features by learning from massive online datasets, pre-trained multi-modal encoders such as CLIP \citep{Radford2021-ro} significantly improved the of quality multi-modal embeddings and provided better input conditioning \citep{Crowson2022-cp}, \citep{Zhou2022-kn}. The visual fidelity of generated images was further improved by the development of Diffusion models \citep{Sohl-Dickstein2015-je}. Diffusion models support the generation of higher-resolution images with fine-grained control over the generation process \citep{Nichol2022-hg}, \citep{Ramesh2021-cu}, \citep{Ding2021-cm}. With the success of language model-based text-only encoders, Large Vision models adopt Large Language Model (LLM) based encoders for T2I generation, leveraging their language comprehension capabilities \citep{Saharia2022-pl}. T2I personalization is another key area of research, where models are trained to generate images of specific concepts and objects with the help of few-shot examples \citep{Kumari2023-uc}, \citep{Ruiz2023-mh}.

\paragraph{Latent Diffusion Models}
Latent Diffusion Models (LDM) are a class of Diffusion models that split the training process into 2 different phases. These models use a compressed latent space for the diffusion process which is perceptually equivalent. An auto-encoder model is leveraged for this purpose, reducing the number of overall trainable parameters. Minimal information loss in the latent space is ensured by training using a patch-based adversarial objective and perceptual loss. The second phase of training improves the conditioning mechanism of Diffusion models by augmenting the UNet with Cross Attention Layers.

\paragraph{Datasets}
In recent literature, there has been a shift towards evaluation-only benchmarks with prompts to judge specific attributes of a generator's performance. PartiPrompts \citep{Saharia2022-pl} and UniBench \citep{Li2022-tf} provide diverse text prompts sorted based on style and difficulty. DiffusionDB \citep{Wang2022-ee} is a large-scale collection of prompt-tuned caption-image pairs commonly used for sourcing captions for T2I evaluation. All these benchmarks contain captions that only provide sparse or detailed descriptions of physical entities within images. We aim to include captions containing situational context information and complex sentence structures as a part of {\n}.

  %-------------------------------------------------------------------------
  \section{Dataset Construction}
  \label{sec:anc-dataset-construction}

  The {\n} (Abstractive News captioNs dAtaset) has been constructed to perform news image generation using abstractive captions. We source images from the NYTimes800K dataset \citep{Tran2020-xl} which contains news articles and associated image-caption pairs scraped from the news organization The New York Times (NYT). This dataset was originally developed for News Image Captioning. Using news image-caption pairs from a reputable media outlet such as NYT helps ensure the dataset's quality. We focus on selecting generalizable entities within our dataset, to observe the impact of different sentence structures and how they affect the generated image. News data contains a multitude of named entities, often with very low repetition and distinct physical appearances, such as faces and geographic landmarks. The inclusion of named entities from news images would drastically increase the complexity of the generative task. The inability to accurately generate named-entity attributes would further hamper context feature representation due to their inter-dependent nature. To combat the mentioned issues, we carefully curate our dataset to include image-caption pairs containing adequate contextual and content-related information. We select samples with lesser dependence on named entities and more general visual components to make the task feasible.      
  
% We still retain captions with commonly observed 'GPE', 'LOC' and 'ORG' entities which weren't tagged in the ground truth annotations.
  \subsection{Pre-processing and Filtering Approaches}

  The original NYTimes800K dataset contains 445K news articles accompanied by 793K image-caption pairs. It spans 14 years of articles published on The NYT website. The dataset has been provided as a MongoDB dump for public access.  The first step of pre-processing focuses on removing image-caption pairs that contain explicit entity mentions both in images and text. We use the provided entity tags for each caption for filtering. We exclude all captions containing the tags 'PERSON', 'GPE', 'LOC', 'WORK\_OF\_ART', and 'ORG'. Some commonly observed 'GPE', 'LOC' and 'ORG' entities as shown in \hyperlink{fig:gen-ex2}{Example 2} were not tagged in the ground truth annotations, which we retain in our dataset. This ensures the removal of obscure visual entities from our dataset. Subsequently, we also set bounds on the caption length between 4 to 70 words. Any captions lesser than 4 words would not be informative enough for extracting usable features and captions larger than 77 words cannot be handled by the CLIP-based Text encoder \citep{Radford2021-ro} that we employ in our experiments. 
  
  Following caption-based filtering, we remove all images where human faces are clearly visible in the foreground. Using a RetinaFace-based face detector \citep{Deng2020-ht}, we remove around 1000 additional images. Through these filtering techniques, we extract relevant image-caption pairs and corresponding article headlines from the NYTimes800K Dataset. All news images in our dataset are uniformly resized to the target input resolution of 512x512. For localization of sections in our source image containing objects of interest, we employ entropy-based region selection \citep{Ma2004-qx}. This makes sure that minimal information loss is present and helps prioritize the foreground objects in each image. \textit{Disclaimer:} The dataset may contain language that is considered profane, vulgar, or offensive by some readers as they are extracted from real-world news articles.
  
  %***----------------------------------------------------------------------

  \subsection{Dataset Insights}
  The filtered and pre-processed version of the {\n} contains 29625 image-text pairs. We split the dataset into Train, Validation, and Testing sets in the ratio of 80\%:10\%:10\% respectively. All metric scores reported have been calculated on the Test set. To better understand the composition of the dataset, we analyze various attributes of the image-text pairs and their source articles. 

  \begin{figure*}
    \centering
    \includegraphics[width=0.8\textwidth]{"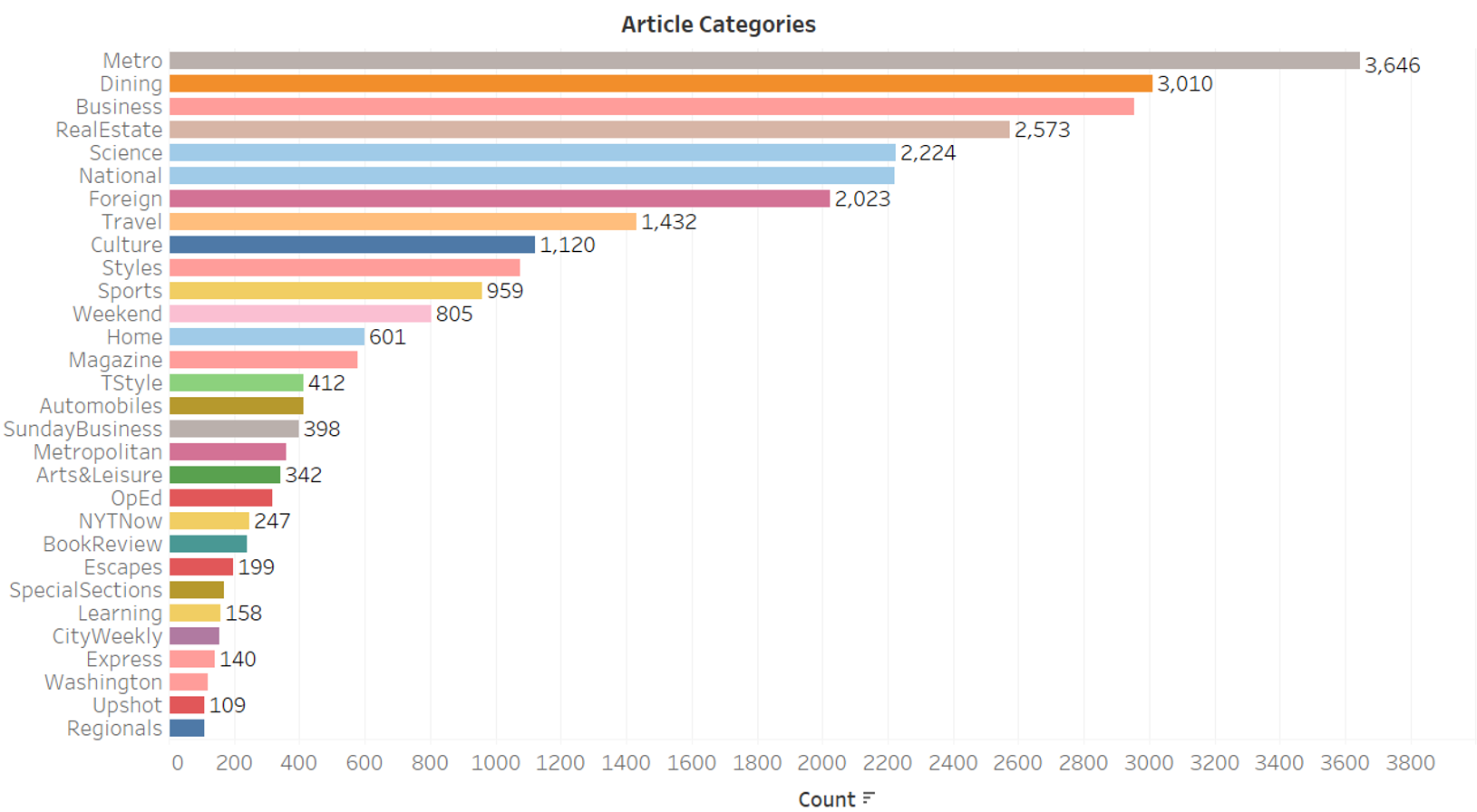"}
    \caption{Visualizing Article Categories of image-caption pairs present in {\n}}
    \label{fig:article_topics}
  \end{figure*}

  \begin{table}[ht]
    \centering
    \resizebox{\linewidth}{!}{
        \begin{tabular}{@{} c|*5c @{}}
        \toprule
        
        \multirow{2}{*}{\bfseries Dataset} & 
        \multirow{2}{*}{\bfseries Unique Tokens} & 
        \multicolumn{2}{c}{\bfseries Caption Length}\\ \cline{3-4}
        && Mean & StdDev \\ \hline
        %------
        COCO Captions Train & 22767 & 10.42 & 0.88 \\
        COCO Captions Val & 16647 & 10.42 & 0.87\\
        CC3M Train & 45896 & 10.31 & 3.30 \\
        CC3M Val & 9289 & 10.40 & 3.35\\
        \hline
        {\n} Train & 17897 & \textbf{14.1} & \textbf{7.75}  \\
        {\n} Validation & 1622 & 13.8 & 7.60 \\
        {\n} Test & 1649 & 14.1 & 7.71 \\
        \bottomrule
        \end{tabular}
    }
       \caption{Dataset Statistics of {\n}}
        \label{table:caption_len}
  \end{table}

  \subsubsection{Caption Statistics}
  In this section, we evaluate different statistical measures for quantifying the distribution of captions across the dataset. Table \ref{table:caption_len} shows the average caption length of captions present in the dataset and across the train, validation, and test subsets. The average caption lengths are similar across the different data splits with the average caption length being slightly greater than that of the COCO Captions dataset. The standard deviation of caption lengths is also higher, indicating the presence of short, long, and multi-line captions. To visualize the token diversity of captions in {\n}, we examine the words by identifying unique token appearances. The unique token count is calculated using the spaCy library, which tokenizes and lemmatizes our captions along and removes all stop words. Subsequently, different Parts of Speech (POS) are tagged and unique tokens belonging to the classes [Common Noun, Proper Noun, Adjectives, and Verbs] are selected. This provides a minimum guarantee that the abstractive captions present are long enough to contain adequate content and contextual features. Our analysis also ensures that the composition of captions present in the train, validation, and test splits are consistent.

  \subsubsection{News Image Analysis}
  Along with the captions, we also estimate image properties such as the number of recognizable objects present in each image and an average number of detected objects per image. We use a YOLO-R-based object detector \citep{Wang2021-bu} for identifying the objects present in each image of our dataset. The YOLO-R detector has been trained on the MS-COCO dataset, containing 80 unique object classes of commonplace objects \citep{Chen2015-qj}. A confidence threshold of 0.4 is used for YOLO-R inference. We find that there are an average of 2.57 objects per image in the {\n}. Fig. \ref{fig:object_freq} shows the most frequently appearing classes of objects in our dataset using a treemap for visualization.

  \subsubsection{Categories of News Articles Selected}
  In this section, we identify the different types of news articles from which image-caption pairs were sourced for dataset construction. In total, there exist 123 unique article topics within our dataset. Only 13 of the image-caption pairs do not have accompanying article type information so we disregard those pairs from our article topic analysis. From Fig. \ref{fig:article_topics}, we see that there exists a good distribution across topics such as Dining, Business, Real Estate, etc. This shows that the news image-caption pairs are diverse and not limited to only a particular type of news article. 

\begin{figure*}
    \centering
    \begin{tabular}{ccccc}
     \textbf{Original} & \textbf{SD 1.5 (Base)} & \textbf{SD 2.1 (Base)} & \textbf{SD 2.1 (LoRA)} & \textbf{SD 2.1 (ReFL)}\\ \toprule
    \includegraphics[width=0.16\textwidth]{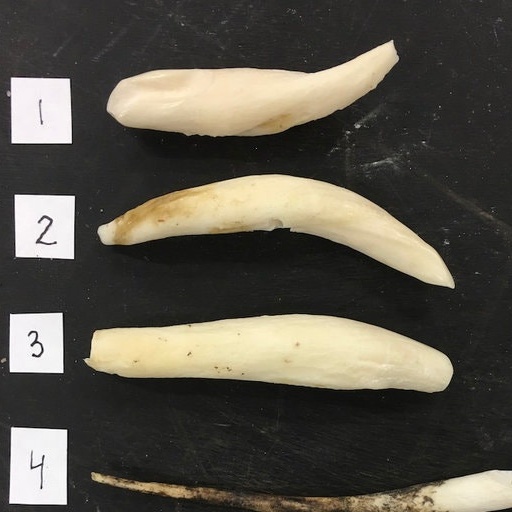} & 
    \includegraphics[width=0.16\textwidth]{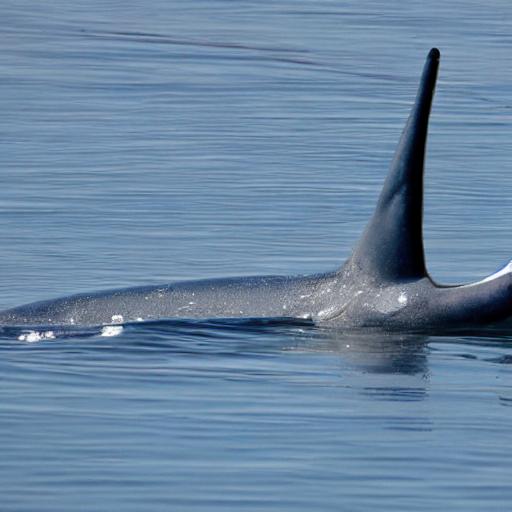} & 
    \includegraphics[width=0.16\textwidth]{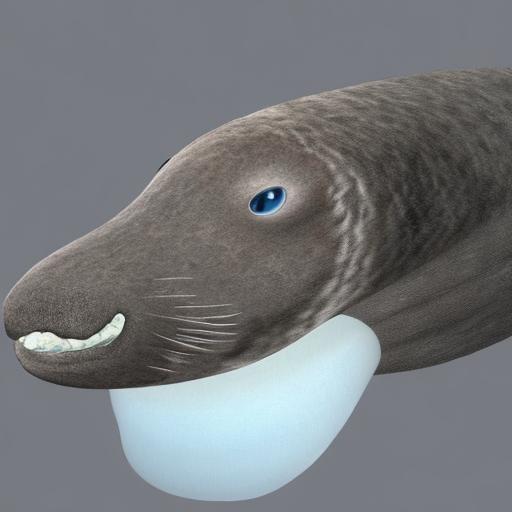} & 
    \includegraphics[width=0.16\textwidth]{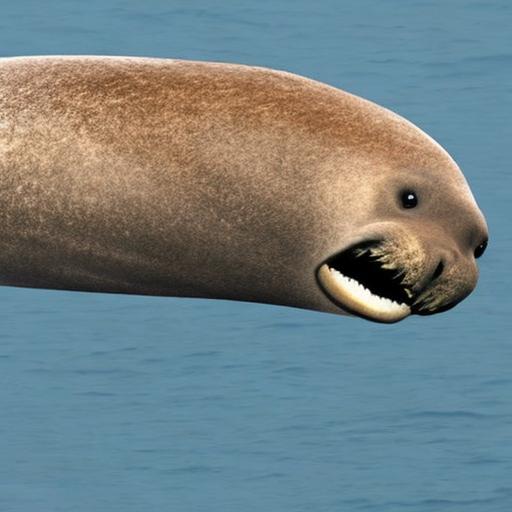} &
    \includegraphics[width=0.16\textwidth]{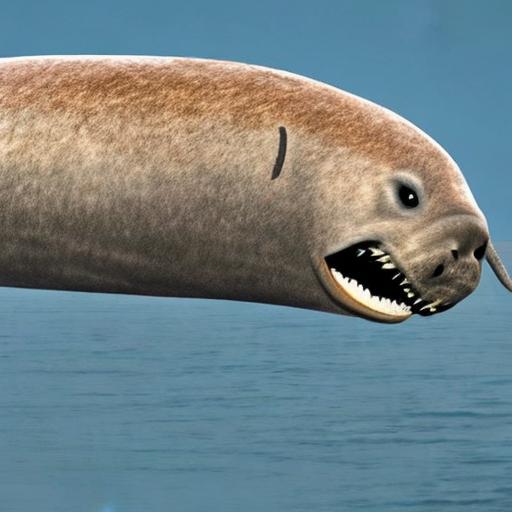} \\
    
    \multicolumn{5}{p{0.85\textwidth}}{\centering \hypertarget{fig:gen-ex1}{\textbf{Caption:} Because of its teeth, the hybrid probably had a diet more like a walrus or bearded seal than a narwhal or a beluga.} } \\ 
    & & & & \\
    
    \includegraphics[width=0.16\textwidth]{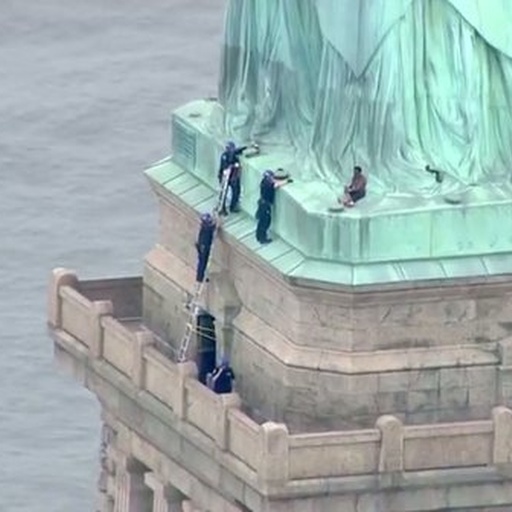} & 
    \includegraphics[width=0.16\textwidth]{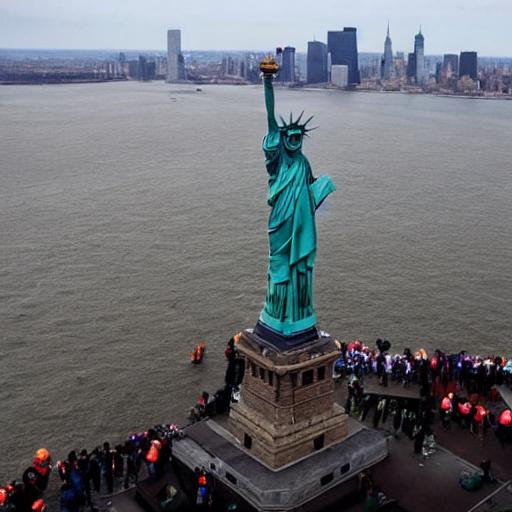} & 
    \includegraphics[width=0.16\textwidth]{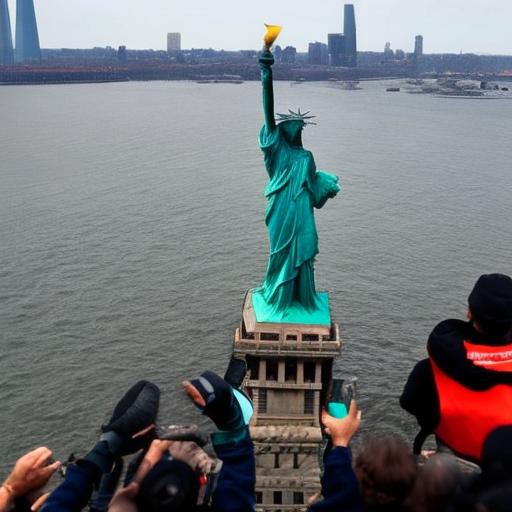} & 
    \includegraphics[width=0.16\textwidth]{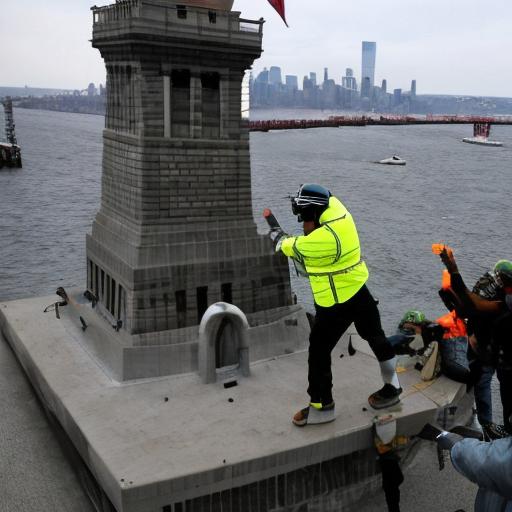} &
    \includegraphics[width=0.16\textwidth]{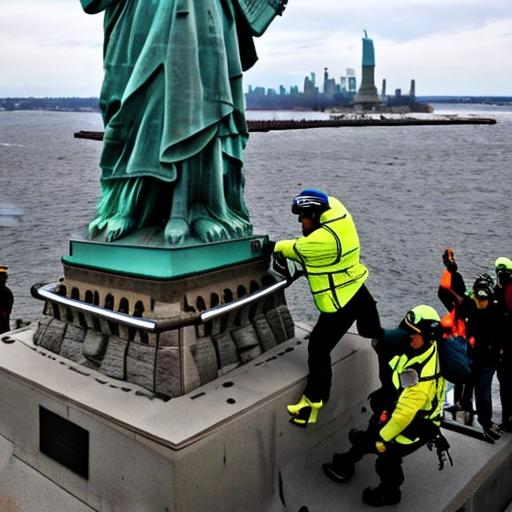} \\
    
    \multicolumn{5}{p{0.85\textwidth}}{\centering  \hypertarget{fig:gen-ex2}{\textbf{Caption:} A protester who scaled the base of the Statue of Liberty forced the shutdown of the monument. The island was cleared of an estimated 4,500 tourists, and the climber was taken safely into custody several hours later.} } \\ 
    & & & & \\

    \includegraphics[width=0.16\textwidth]{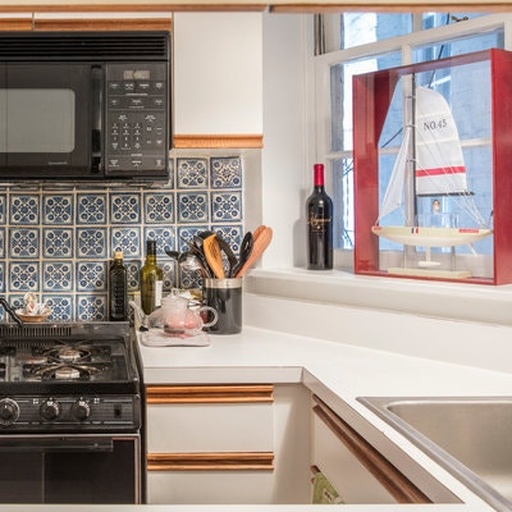} & 
    \includegraphics[width=0.16\textwidth]{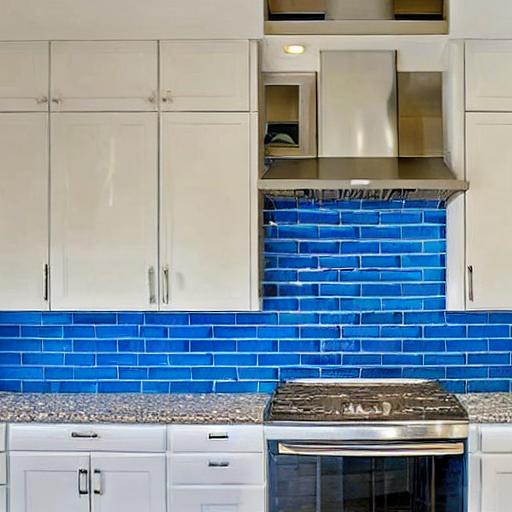} & 
    \includegraphics[width=0.16\textwidth]{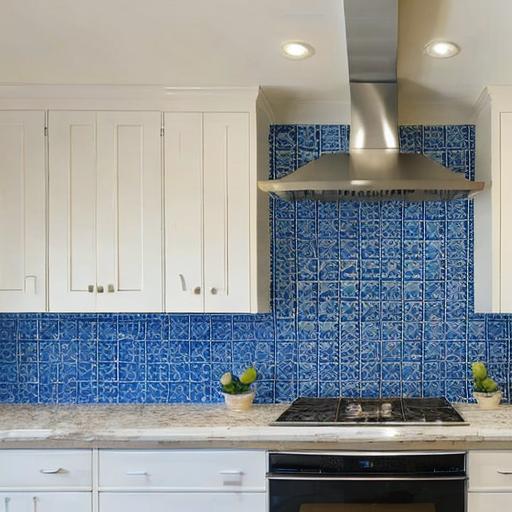} & 
    \includegraphics[width=0.16\textwidth]{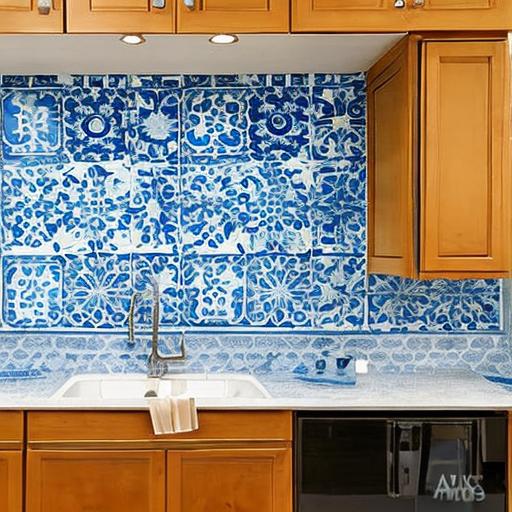} &
    \includegraphics[width=0.16\textwidth]{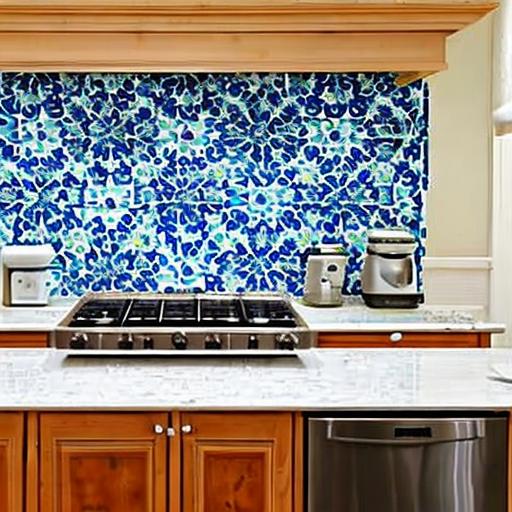} \\
    
    \multicolumn{5}{p{0.85\textwidth}}{\centering \hypertarget{fig:gen-ex3}{\textbf{Caption:} The kitchen has a pretty blue-and-white tile backsplash.} } \\ 
    & & & & \\

    \includegraphics[width=0.16\textwidth]{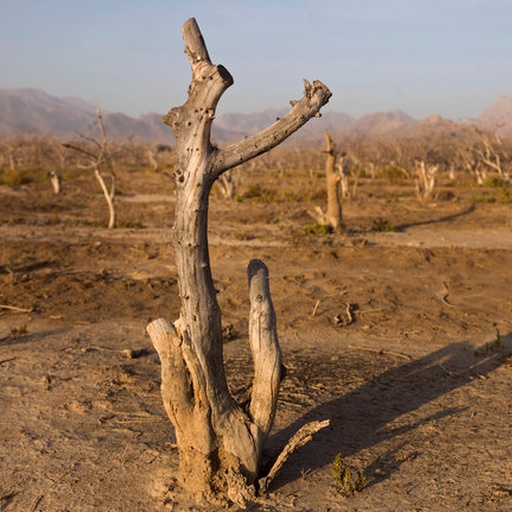} & 
    \includegraphics[width=0.16\textwidth]{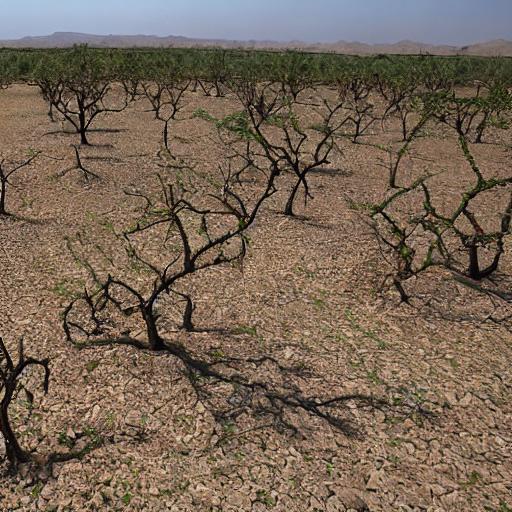} & 
    \includegraphics[width=0.16\textwidth]{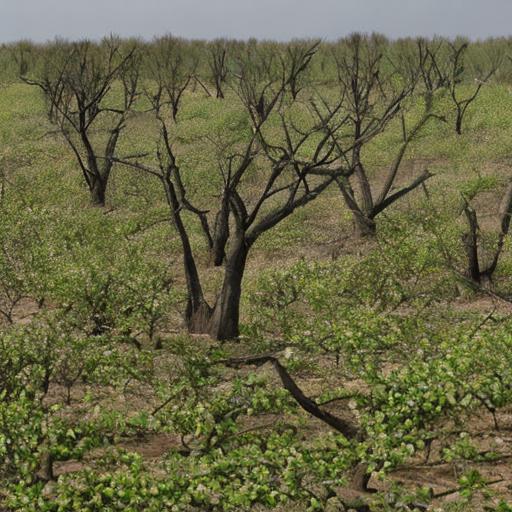} & 
    \includegraphics[width=0.16\textwidth]{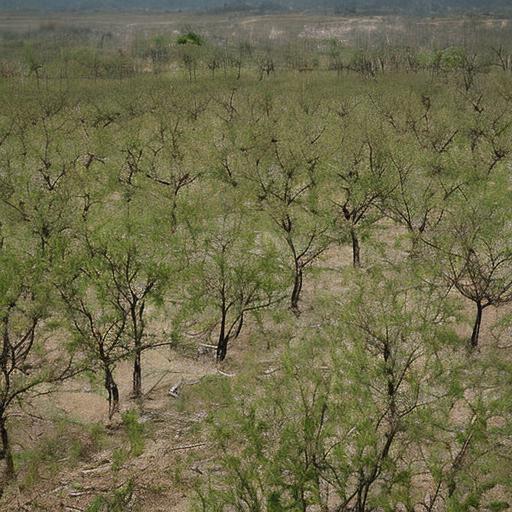} &
    \includegraphics[width=0.16\textwidth]{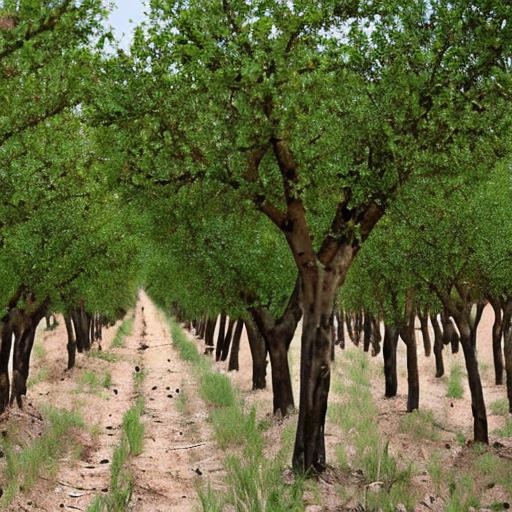} \\
    
    \multicolumn{5}{p{0.85\textwidth}}{\centering  \hypertarget{fig:gen-ex4}{\textbf{Caption:} An orchard of dead trees near the village of Pouze Khoon. Until a decade ago, the pistachio groves were green. Now there is no rain and the groundwater is almost all gone.} } \\ 
    & & & & \\

    \includegraphics[width=0.16\textwidth]{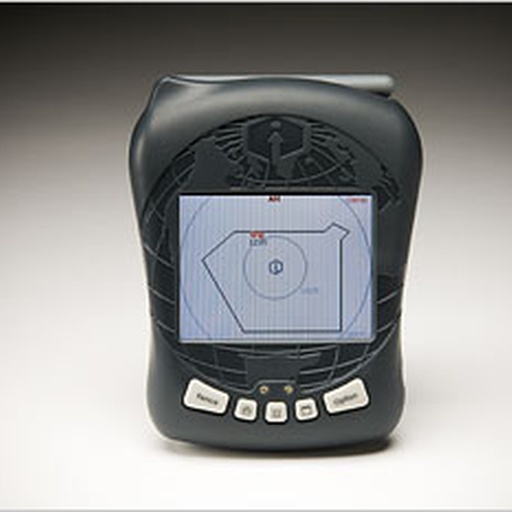} & 
    \includegraphics[width=0.16\textwidth]{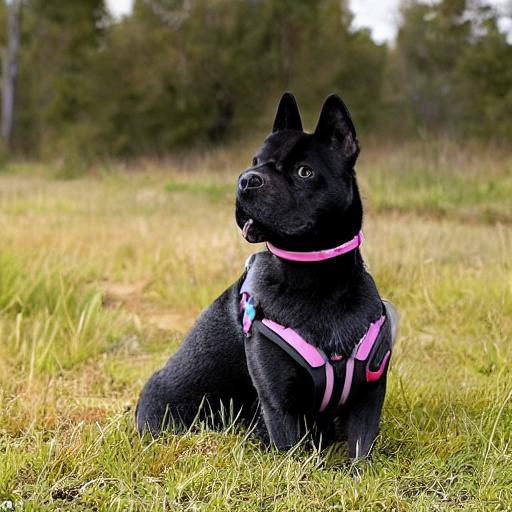} & 
    \includegraphics[width=0.16\textwidth]{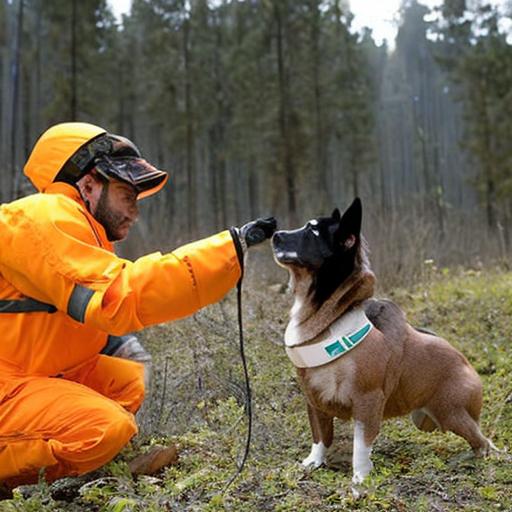} & 
    \includegraphics[width=0.16\textwidth]{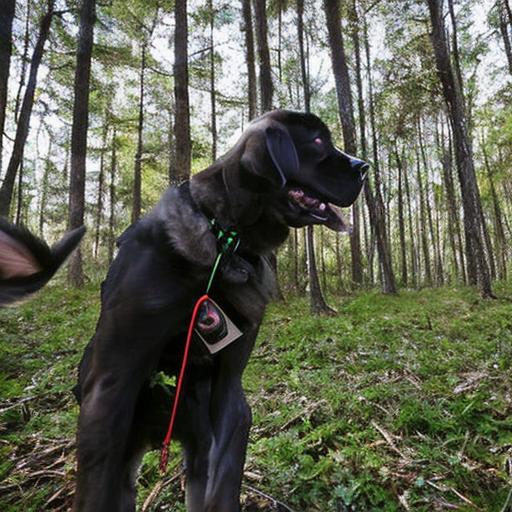} &
    \includegraphics[width=0.16\textwidth]{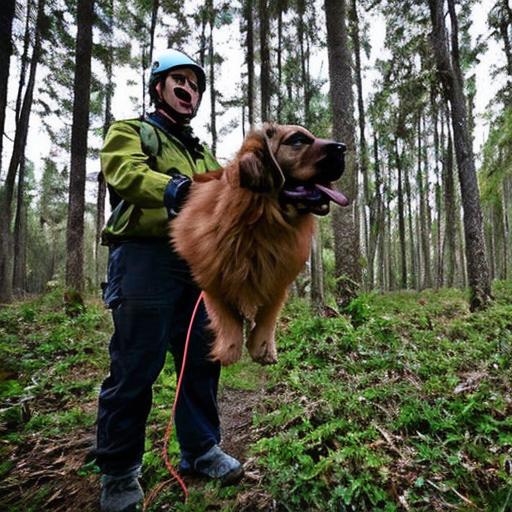} \\
    
    \multicolumn{5}{p{0.85\textwidth}}{\centering \hypertarget{fig:gen-ex5}{\textbf{Caption:} With the RoamEO base unit, left (which includes a collar), a dog owner can get radio signals tracking the animal's location, up to 1.5 miles away.} } \\ 
    & & & & \\

  \end{tabular}
  \label{figure:gen_results}
\end{figure*}

\begin{figure*}
\centering
\begin{tabular}{ccccc}
  \textbf{Original} & \textbf{SD 1.5 (Base)} & \textbf{SD 2.1 (Base)} & \textbf{SD 2.1 (LoRA)} & \textbf{SD 2.1 (ReFL)}\\ \toprule

    \includegraphics[width=0.16\textwidth]{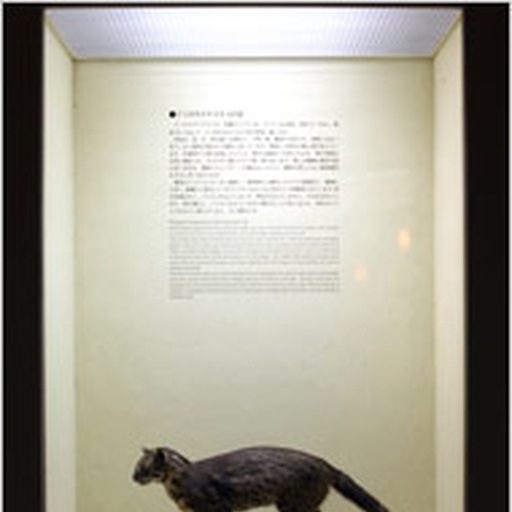} & 
    \includegraphics[width=0.16\textwidth]{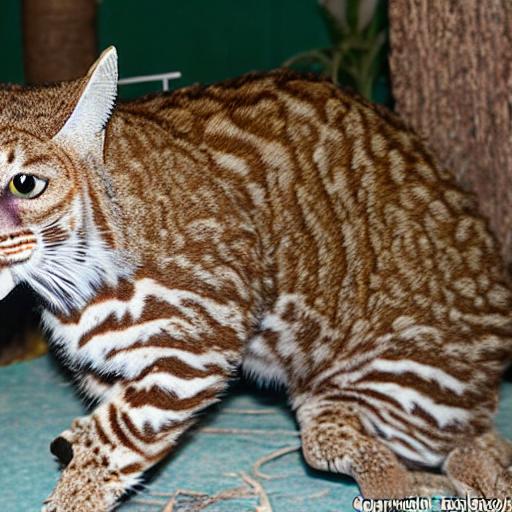} & 
    \includegraphics[width=0.16\textwidth]{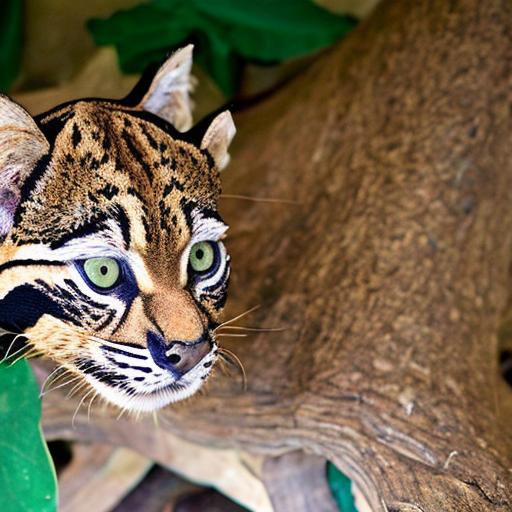} & 
    \includegraphics[width=0.16\textwidth]{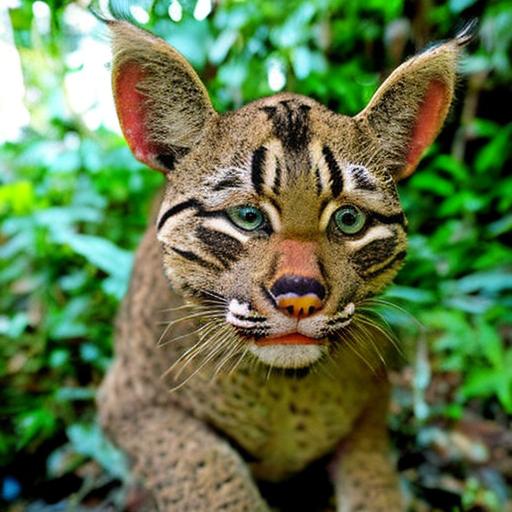} &
    \includegraphics[width=0.16\textwidth]{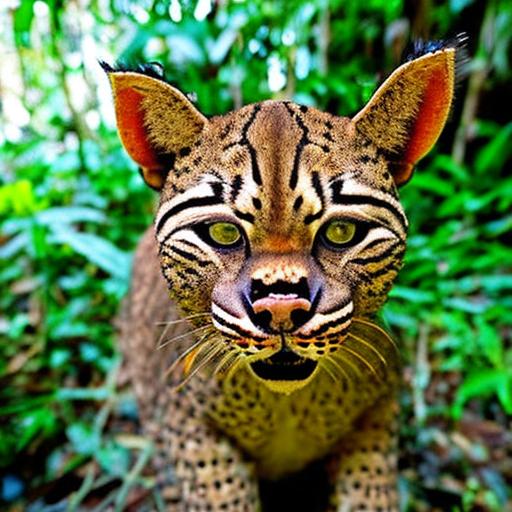} \\
    
    \multicolumn{5}{p{0.85\textwidth}}{\centering \hypertarget{fig:gen-ex6}{\textbf{Caption:} Unlike this stuffed Iriomote wildcat at a wildlife conservation center, live specimens are elusive and seldom seen.} } \\

    \end{tabular}
        \caption{Qualitative comparison of different T2I models on {\n} Dataset}
        \label{figure:gen_results}
    \end{figure*}

    \begin{table*}[!htbp]
      \centering
      \resizebox{0.65\linewidth}{!}{
          \begin{tabular}{@{} c|*3c @{}}
          \toprule
          \large{Model} & \large{$FID_{CLIP}$} (↓) & \large{ImageReward (↑)} & \large{HPS V2 (↑)}\\ 
          % \hline
          \midrule
          \midrule
  
          Stable Diffusion 2.1 (ReFL) & 9.728 & \textbf{0.2182} & \textbf{0.2470}\\
  
          Stable Diffusion 2.1 (LoRA) & \textbf{7.5906} & -0.0081 & 0.2335\\
      
          Stable Diffusion 2.1 (Base) & 7.9707 & 0.1041 & 0.2399\\
          % \hline
          Stable Diffusion 1.5 (Base) & 7.7008 & -0.0094 & 0.2312\\
          % \hline
          Stable Diffusion 1.4 (Base) & 7.7178 & -0.0104 & 0.2314\\
  
          Lafite (Fine-tuned)         & 12.5602 & -0.933 & 0.1809\\
  
          Lafite (Base)               & 20.8314 & -1.3477 & 0.1685\\
          
          \bottomrule
          \end{tabular}
          }
  
      \caption{Results of Abstractive T2I synthesis on {\n}}
      \label{table:metrics}
    \end{table*}

  \section{Evaluation Methodology}
    \label{sec:evaluation-methodology}
    
    To understand how different architectures learn abstractive captions on the {\n}, we consider various T2I synthesis models previously proposed in the literature. We select 2 major architectures: Latent Diffusion Models and StyleGAN-based generators for this task. As all these models have achieved State-of-the-Art scores on descriptive caption datasets, we evaluate how they perform with news domain-specific, abstractive captions in our experiments and visualize the results. All images are generated in 512x512 resolution.

    \paragraph*{T2I Synthesis Models}
    We select SOTA T2I models: Stable Diffusion (SD) V1.4, V1.5, and V2.1 \citep{Rombach2022-ci} as baselines for the task of news image generation. They belong to the class of Latent Diffusion Models (LDM), which have shown strong performance in traditional benchmarks such as COCO Captions \citep{Chen2015-qj}. We also evaluate Lafite \citep{Zhou2022-kn}, which utilizes a pre-trained CLIP encoder for translating text embeddings into the image feature space. It adapts an unconditional StyleGAN2 generator \citep{Karras2020-or} by injecting text-conditional information through affine transformations. Two Fully Connected Layers are utilized to transform the input text features to be more semantically similar to StyleGAN's image Stylespace. 

    \paragraph*{Fine-tuning Approaches}
    We test 2 parameter-efficient fine-tuning approaches for SD V2.1 Low-Rank Adaptation (LoRA) \citep{Hu2021-ft} and Reward Feedback Learning (ReFL) \citep{Xu2023-rj} on the {\n} dataset. Instead of modifying the model weights directly during the fine-tuning process, LoRA injects trainable rank decomposition matrices for each transformer layer within the model. This allows for faster convergence and better generalization on smaller datasets. ReFL, on the other hand, employs reward models for providing feedback to the generator instead of directly fine-tuning on the dataset itself. These reward models are trained to predict human preference scores for a provided image-caption pair, acting as an indirect indicator for image-caption alignment. ReFL only utilizes captions for tuning the generator and does not require additional image data.

    \paragraph*{Metrics} 
    To holistically evaluate the samples generated by T2I models on {\n}, we report 3 different types of metrics: Frechet Inception Distance (FID) \citep{Heusel2017-ob}, ImageReward \citep{Xu2023-rj} and Human Preference Score (HPS) V2  \citep{Wu2023-nm}. FID serves as an indicator to quantify the overall realism and diversity of generated samples compared to the ground truth images. With the distribution of datasets like {\n} diverging significantly from the Inception-V3 used in traditional FID calculations \citep{Kynkaanniemi2022-de}, we adopt the more representative $FID_{CLIP}$ metric for our testing. To measure the relatedness of our generated images and ground truth captions, we utilize ImageReward. Compared to image-caption similarity metrics like CLIPScore \citep{Hessel2021-we}, ImageReward is trained on real-world image-caption pairs annotated and ranked according to human preference. Similarly, HPS V2 also serves as an indicator of preference alignment.

  \section{Experiments}
    \label{sec:experiments}
  \paragraph*{Implementation Details}
  In our experiments, we train Lafite on {\n} in a fully-supervised setting. We train 2 variants of Lafite, with and without fine-tuning. In the base variant, we train it from scratch on the {\n} until convergence for 4400 epochs. To perform fine-tuning, we initialize the model with pre-trained weights from the Conceptual Captions (CC3M) dataset \citep{Sharma2018-tr} and continue training on {\n} till convergence for 1800 epochs. Both Lafite models use the hyperparameters: \textit{itd} = 5, \textit{itc} = 10, \textit{gamma} = 10 and \textit{temp} = 0.5. For the Stable Diffusion models, we set \textit{guidance\_scale=7.5},  \textit{mixed\_precision=fp16}, and \textit{inference\_steps=100}. We set \textit{max\_grad\_norm=1} and perform LoRA fine-tuning for 1000 epochs. Similarly for ReFL, we set \textit{max\_grad\_norm=2}, \textit{grad\_scale=0.001} and fine-tune for 200 epochs. Both methods used \textit{batch\_size=2}, \textit{grad\_accumulation\_steps=4}, \textit{use\_ema=True}, \textit{learning\_rate=1e-05} and \textit{lr\_scheduler=constant}. All reported metrics are averaged across 2 random seeds 42 and 973. Table \ref{table:metrics} shows the quantitative results while Fig. \ref{figure:gen_results} presents a qualitative comparison of our experiments on the {\n} dataset.

  \subsection{Quantitative Results}
  Looking at the results in Table \ref{table:metrics}, we observe that different types of fine-tuned Stable Diffusion models outperform their zero-shot counterparts on specific metrics. This supports our hypothesis that pre-training datasets contain a majority of descriptive prompts rather than image captions. From the reported $FID_{CLIP}$ scores, we observe that Stable Diffusion (LoRA) scores the best in terms of producing high-fidelity samples that are well aligned with the ground truth image distribution. On the other hand, Stable Diffusion (ReFL) performs the best in terms of ImageReward and Human Preference Score V2. This indicates that image quality is not directly correlated with image-caption relevance and human preference. Traditional T2I models only utilize image quality metrics for quantifying model performance which may not be the best indicator for abstractive caption datasets.

  \subsection{Qualitative Results}
  From the visualizations in Figure \ref{figure:gen_results}, we observe that the fine-tuned Stable Diffusion models capture key subjects mentioned in the caption accurately. In \hyperlink{fig:gen-ex1}{Example 1}, the caption primarily describes the teeth of an animal. Stable Diffusion 2.1 (ReFL) showcases its improved syntactic awareness compared to all other models by presenting a more accurate representation of the animal's teeth. Similarly, in \hyperlink{fig:gen-ex2}{Example 2}, the non-fine-tuned model fails to capture the climber's presence in the image. Additionally, they do not understand the contextual cues of the caption and show multiple spectators present near the Statue of Liberty when the caption specifically states they were cleared. Only Stable Diffusion 2.1 (ReFL) manages to depict the statue and the climber in adequate detail. This shows that fine-tuning with abstractive captions can significantly improve the model's understanding of context and content features.  However, even the fine-tuned models fail to capture the key subjects in a caption in certain scenarios. In \hyperlink{fig:gen-ex4}{Example 4}, the caption describes a forest of dead trees as the main subject of the image. All the models instead generate forests with trees that still have green leaves. With \hyperlink{fig:gen-ex6}{Example 6}, all models failed to generate a stuffed animal and instead, generated a real wildcat. This issue arises when an object of importance needs to be generated in an atypical context not commonly observed. By testing these models on more abstract captions, we can better understand the limitations of current T2I models and develop strategies to improve their performance consistently.

  %------------------------------------------------------------------------
\section{Discussion and Conclusion}
\label{sec:discussion-and-conclusion}

  Our experiments demonstrate how existing open-source T2I architectures comprehend abstractive captions present in domain-specific data such as news media. We show that focusing on either image quality or image-caption relevance is not sufficient when evaluating T2I synthesis on more complicated captions. Our proposed dataset {\n} helps facilitate the development of more linguistically aware T2I models capable of integrating reasoning and context cues in the generation process. Although existing fine-tuning approaches show promise in improving model performance, they still struggle to jointly optimize both image fidelity and human preference alignment in generated images. This highlights the need for more sophisticated architectures that can understand the nuances of different feature types present in abstractive captions. As the size of datasets keeps increasing, scaling up human annotation of images to match demand adds a huge overhead. Descriptive prompts need to be tightly coupled with the reference image's contents, needing multiple rounds of evaluation and filtering, which makes it a manually tedious task. The use of abstractive captions for images can greatly simplify the human annotation process for datasets. Additionally, {\n} motivates the development of journalism assistance solutions. The use of keywords and descriptive prompts with current image generators involves a lot of prompt engineering to get relevant images for a specific topic \citep{Liu2022-ut}. High-quality images are generated only when a particularly restrictive sentence structure and vocabulary are used. With models trained to understand abstractive captions, the requirements for intensive prompt engineering would be significantly reduced. Similarly, achieving better delineation between different feature types present in non-descriptive prompts can also benefit related tasks such as image retrieval. The addition of context can play a major role in influencing the quality of retrievals.

  \paragraph*{Limitations}
  This paper aims to introduce the potential of abstractive captions to motivate the development of more contextually grounded T2I synthesis models, particularly when synthesizing news-domain-specific images. Although news articles contain a lot of named entities, we choose to filter them out and instead focus on context features that can be inferred from text captions and depicted by general visual concepts. Developing T2I synthesis architectures that can take advantage of named entities using external knowledge bases as references would help overcome this limitation. Another challenge in our evaluation methodology is the max prompt length restriction for multi-modal encoders such as CLIP. With an input prompt length of 77 tokens, we are constrained in terms of caption length for which we can effectively evaluate the task of news image generation. To tackle this challenge, we plan to experiment with alternate LLM-based multi-modal encoders capable of accepting longer abstractive captions. From the perspective of metric selection, large-scale human evaluation of images generated by T2I architectures on abstractive captions is an important step toward measuring their relative performance. We aim to run human evaluation studies as a part of our future research.
  
  \paragraph*{Potential negative societal impacts}
  Image generation architectures have the potential to be misused for nefarious use cases such as spreading disinformation \citep{Zhang2022-ix} and generating neural fake news \citep{Zellers2019-dn}. Our current pre-processing pipeline removes most images containing named entities, i.e., public figures and locations of national importance, contributing to risk mitigation. However, we recognize the threat posed by contextually relevant Deepfake images when dealing with news media images. Future research directions include understanding the extent to which T2I models can be used for neural fake news generation and identifying appropriate detection strategies.

\section*{Acknowledgments}
This research was in part supported by the U.S. National Science Foundation (NSF) award \#1820609. Part of the research results were obtained using the computational resources provided by CloudBank (\url{https://www.cloudbank.org/}), which was supported by the NSF award \#1925001.

\bibliography{papers}

\end{document}